\documentclass{colt2020} 

\title[Data-driven confidence bands for distributed nonparametric regression]{Data-driven confidence bands for distributed nonparametric regression}
\usepackage{times}

\RequirePackage[OT1]{fontenc}
\RequirePackage{amsmath}

\usepackage{amsmath}
\usepackage{mathtools}

\mathtoolsset{showonlyrefs}
\usepackage{amssymb}
\usepackage{calligra}

\coltauthor{%
 \Name{Valeriy Avanesov} \Email{avanesov@wias-berlin.de}\\
 \addr WIAS Berlin
}

\makeatletter
\def\renewtheorem#1{%
	\expandafter\let\csname#1\endcsname\relax
	\expandafter\let\csname c@#1\endcsname\relax
	\gdef\renewtheorem@envname{#1}
	\renewtheorem@secpar
}
\def\renewtheorem@secpar{\@ifnextchar[{\renewtheorem@numberedlike}{\renewtheorem@nonumberedlike}}
\def\renewtheorem@numberedlike[#1]#2{\newtheorem{\renewtheorem@envname}[#1]{#2}}
\def\renewtheorem@nonumberedlike#1{  
	\def\renewtheorem@caption{#1}
	\edef\renewtheorem@nowithin{\noexpand\newtheorem{\renewtheorem@envname}{\renewtheorem@caption}}
	\renewtheorem@thirdpar
}
\def\renewtheorem@thirdpar{\@ifnextchar[{\renewtheorem@within}{\renewtheorem@nowithin}}
\def\renewtheorem@within[#1]{\renewtheorem@nowithin[#1]}
\makeatother

\newtheorem{assumption}{}
\renewtheorem{theorem}{}
\renewtheorem{lemma}{}
\renewtheorem{corollary}{}
\renewtheorem{remark}{}

\begin{document}

\renewcommand{\thedefinition}{Definition \thesection.\arabic{definition}}
\renewcommand{\thelemma}{Lemma \arabic{lemma}}
\renewcommand{\thetheorem}{Theorem \arabic{theorem}}
\renewcommand{\theassumption}{Assumption \arabic{assumption}}
\renewcommand{\theremark}{Remark \arabic{remark}}
\renewcommand{\thecorollary}{Corollary \arabic{corollary}}

\maketitle
\newcommand{\fhat}{\hat f}
\newcommand{\fhatm}{\hat f_p}
\newcommand{\fhatmb}{\hat f_p^\flat}
\newcommand{\fbar}{\bar f}
\newcommand{\fbarb}{\bar f^\flat}
\newcommand{\Hknorm}[1]{\left\|#1\right\|_{\mathcal{H}_k}}
\newcommand{\Hsnullnorm}[1]{\left\|#1\right\|_{\mathcal{H}^{s_\circ}}}
\newcommand{\hatSigma}{\hat \Sigma}
\newcommand{\Sigmastar}{\Sigma^*}
\newcommand{\snull}{s_\circ}
\newcommand{\Lrho}{L_{\rho}}
\newcommand{\normSq}[1]{\left\|#1\right\|^2}
\newcommand{\norm}[1]{\left\|#1\right\|}
\newcommand{\normF}[1]{\left\|#1\right\|_F}
\newcommand{\normOne}[1]{\left\|#1\right\|_1}
\newcommand{\fstar}{f^*}
\newcommand{\fstarbold}{\boldsymbol{\mathrm{f}}^*}
\newcommand{\R}{\mathbb{R}}
\newcommand{\gboldm}{\boldsymbol{\mathrm{g}}(p)}
\newcommand{\gboldbm}{\boldsymbol{\mathrm{g}}^\flat(p)}
\newcommand{\brac}[1]{\left(#1\right)}
\newcommand{\cbrac}[1]{\left\{#1\right\}}
\renewcommand{\H}{\mathcal{H}_k}
\newcommand{\Hsnull}{\mathcal{H}^{s_\circ}}
\newcommand{\Hs}{\mathcal{H}^{s}}
\newcommand{\f}{\mathbf{f}}
\newcommand{\fhatbold}{\mathbf{\hat f}}
\newcommand{\fhatboldm}{\mathbf{\hat f}(p)}
\newcommand{\fhatboldmb}{\mathbf{\hat f}^\flat(p)}
\newcommand{\fbarbold}{\mathbf{\bar f}}
\newcommand{\fbarboldb}{\mathbf{\bar f}^\flat}
\newcommand{\ftruebold}{\mathbf{f^*}}
\newcommand{\ftrueboldrho}{\mathbf{f}^*_{\rho}}
\newcommand{\ftruerho}{f^*_{\rho}}
\newcommand{\dotprod}[2]{\langle#1,#2\rangle}
\newcommand{\param}{\boldsymbol{\theta}}
\newcommand{\gdelta}{\boldsymbol{\gamma}_{\delta}}
\newcommand{\ghat}{\boldsymbol{\hat \gamma}}
\newcommand{\gbar}{\boldsymbol{\bar \gamma}}
\newcommand{\ggamma}{\boldsymbol{\gamma}}
\newcommand{\ggammab}{\boldsymbol{\gamma}^\flat}
\newcommand{\trueparam}{\boldsymbol{\theta^*}}
\newcommand{\trueparamrho}{\boldsymbol{\theta^*_\rho}}
\newcommand{\hatparam}{\boldsymbol{\hat \theta}}
\newcommand{\hatparamm}{\boldsymbol{\hat \theta}_p}
\newcommand{\E}[1]{\mathbb{E}\left[#1\right]}
\newcommand{\empE}[1]{\mathbb{E_D}\left[#1\right]}
\newcommand{\Eb}[1]{\mathbb{E}^{\flat}\left[#1\right]}
\newcommand{\Var}[1]{\mathrm{Var}\left[#1\right]}
\newcommand{\score}{\nabla \zeta}
\newcommand{\Psit}{\Psi^T}
\newcommand{\Psimt}{\Psi^T_p}
\newcommand{\Psim}{\Psi_p}
\newcommand{\eps}{\boldsymbol{\varepsilon}}
\newcommand{\epst}{\boldsymbol{\varepsilon}^T}
\newcommand{\epsm}{\boldsymbol{\varepsilon}(p)}
\newcommand{\epsmt}{\boldsymbol{\varepsilon}(p)^T}
\newcommand{\Drho}{D_\rho}
\newcommand{\DrhoSq}{D_\rho^2}
\newcommand{\DrhoSqInv}{D_\rho^{-2}}
\newcommand{\Drhom}{D_\rho(p)}
\newcommand{\DrhomSq}{D_\rho^2(p)}
\newcommand{\DrhomSqInv}{D_\rho^{-2}(p)}
\newcommand{\Prob}[1]{\mathbb{P} \left\{#1\right\}}
\newcommand{\Probb}[1]{\mathbb{P}^\flat \left\{#1\right\}}
\newcommand{\DrhoInv}{D_\rho^{-1}}
\newcommand{\DrhomInv}{D_\rho^{-1}(p)}
\newcommand{\barparam}{\boldsymbol{\bar \theta}}
\newcommand{\inv}[1]{#1^{-1}}
\newcommand{\MInv}{\inv{M}}
\newcommand{\phiinf}{C_\phi}
\newcommand{\C}{{\mathrm{C}}}
\let\oldc=\c
\renewcommand{\c}{{\mathrm{c}}}
\newcommand{\Rclt}{\mathrm{R_{CLT}}}
\newcommand{\Rb}{\mathrm{R^{\flat}}}
\newcommand{\tr}[1]{\mathrm{tr}\brac{#1}}
\newcommand{\g}{\mathfrak{g}}
\newcommand{\x}{\mathrm{x}}
\newcommand{\p}{\mathfrak{p}}
\renewcommand{\t}{\mathrm{t}}
\renewcommand{\u}{\mathrm{u}}
\renewcommand{\r}{\mathrm{r}}
\newcommand{\X}{\mathcal{X}}
\newcommand{\matern}{Matérn}
\newcommand{\kdd}{k(\cdot, \cdot)}
\newcommand{\abs}[1]{\left|#1\right|}
\newcommand{\minusroot}{^{-1/2}}
\newcommand{\proot}{^{1/2}}
\newcommand{\Frho}{\brac{M^2 + \rho I}}
\newcommand{\indicator}[1]{\mathbb{I}[#1]}
\newcommand{\given}{\middle|}
	
\begin{abstract}%
Gaussian Process Regression and Kernel Ridge Regression are popular nonparametric regression approaches. 
Unfortunately, they suffer from high computational complexity rendering them inapplicable to the modern massive datasets. 
To that end a number of approximations have been suggested, some of them allowing for a distributed implementation. 
One of them is the divide and conquer approach, splitting the data into a number of partitions, obtaining the local estimates and finally averaging them. 
In this paper we suggest a novel computationally efficient fully data-driven algorithm, quantifying uncertainty of this method, yielding frequentist $L_2$-confidence bands.
We rigorously demonstrate validity of the algorithm. 
Another contribution of the paper is a minimax-optimal high-probability bound for the averaged estimator, complementing and generalizing the known risk bounds. 
\end{abstract}


\section{Introduction}\label{secintro}
The problem of nonparametric regression arises in numerous applications including but not limited to finance \citep{degl2017bank,xu2017nonparametric,tzeremes2018financial,basu2016interpolating}, neuroimaging \citep{hyun2014sgpp,reiss2014massively}, climate \citep{honda2014heat,perez2016features,mudelsee2019trend}, geology \citep{di2014nonparametric,lary2016machine,kirkwood2015geological} and optimization \citep{Srinivas10gaussianprocess,Chowdhury2017}. 
The frequentist setting of such a problem considers $n$ response-covariate pairs $(y_i, X_i)$ from $\R\times \X$ are being observed such that 
\begin{equation}
  y_i = \fstar(X_i) + \varepsilon_i
\end{equation}
for a compact $\X\subseteq \R^d$, centered independent sub-Gaussian noise  $\varepsilon_i$ of variance $\sigma^2$. Throughout the paper we presume $X_i$ are drawn independently w.r.t. some  unknown continuous measure $\pi$. 
In the paper we investigate the behaviour of one of the most popular non-parametric approaches to estimation of $\fstar$ -- the Gaussian Process Regression (GPR) \citep{RW,bilionis2012multi,he2011single,chen2013gaussian}. 
GPR, being a Bayesian procedure, has been predominantly examined from a Bayesian point of view, i.e. no existence of $\fstar$ has been presumed, contrary to the frequentist setting we consider here. Namely, the contraction rate of the posterior distribution has been typically in focus \citep{vaart2011information,van2009adaptive,bhattacharya2017posterior}. 

Commonly, in order to analyse the frequentist behaviour of GPR, researchers turn to the Kernel Ridge Regression (KRR), whose point estimate $\fhat$ coincides with the mean of GPR posterior \citep{Caponnetto2007,mendelson2002geometric,van2006empirical,koltchinskii2006local,zhang2005learning}. 
Most recently researchers have also developed interest for frequentist confidence sets. Namely, the authors of \citep{Yang2017} suggest an approach to construct sup-norm confidence bands. 

Unfortunately, KRR and GPR suffer from $O(n^3)$ time complexity, which renders them inapplicable to the datasets containing more than several thousands elements. To that end numerous approximations emerged. A wide variety of them rely on a low-rank approximation of the kernel driving the GP prior, e.g. PCA \citep{scholkopf1998nonlinear} or Nyström approximation \citep{williams2001using}. The latter has been demonstrated to achieve nearly minimax-optimal performance  \citep{bach2013sharp,fine2001efficient,Rudi2015}. 
Another strategy is to split the dataset into $P$  partitions, obtain the local estimates $\fhatm$ separately and then average them, yielding $\fbar$. 
This does not only reduce the complexity to $O(n^3/P^2)$, but also makes a distributed implementation trivial. 
In \citep{Rosenblatt2016} the method is given theoretical treatment for parametric families of dimenionality $p(n)/n \rightarrow{ const} \in (0,1)$.
 \citep{Deisenroth2015} employed the idea for GPR, in \citep{Zhang2015} the idea is applied to KRR and further studied in \citep{lin2017distributed}. A broad range of distributed non-parametric methods is analysed in \cite{8bdaab8de3cc42f7b13b66b5a1c01f3d}. 

One of the properties making GPR the instrument of choice is the ability to quantify the uncertainty of prediction. 
Only recently \citep{Yang2017} have demonstrated that GPR posterior can be used to construct sup-norm frequentist confidence bands. 
At the same time, data driven-techniques based on Kernel Density Estimator were suggested \citep{Hall2013,Cheng2019}.
To the best of the authors' knowledge there has not yet been a distributed approach to the problem of nonparametric regression yielding frequentist uncertainty estimates. 

The main contribution of this paper is enrichment of the divide and conquer approach suggested by \cite{Zhang2015} with a highly cost-effective novel bootstrap algorithm constructing confidence bands for $\fstar$. 
We rigorously demonstrate the validity of the approach.
The main result is established for undersmoothed prior, which is an assumption commonly employed to establish the validity of confidence sets \citep{knapik2011bayesian,szabo2015frequentist,ray2017adaptive,Yang2017}. 
Moreover, we also obtain a minimax-optimal high-probability bound for $\normSq{\fbar - \fstar}_2$, extending the earlier results \citep{Zhang2015,lin2017distributed,Mucke2018}, where the authors control expectation of the norm or of its positive power.

\subsection{Notation}
In the paper we heavily rely on a spectral decomposition of the kernel operator $\kdd$ w.r.t. $\pi$.  Mercer's theorem \citep{RW} provides existence of normalized eigenfunctions $\phi_j \in L_2(\X, \pi)$ along with the corresponding eigenvalues $\mu_j$ (in decreasing order). For a function $\norm{\cdot}_2$ denotes an $L_2(\X, \pi)$-norm, namely $\norm{f}_2^2 = \int f^2 d\pi$, the dot-product is also defined w.r.t. $\pi$: $\dotprod{f}{g} = \int fgd\pi$. 
The kernel $\kdd$ induces a RKHS $\H$ endowed with a norm 
\begin{equation}\label{rkhsnormdef}
  \norm{f}^2_{\H} = \sum_{j=1}^{\infty} \frac{\dotprod{f}{\phi_j}^2}{\mu_j}.
\end{equation}
For a vector $\norm{\cdot}$ denotes an $\ell_2$-norm, while for a matrix it denotes its maximum absolute eigenvalue. Frobenius norm is denoted as $\norm{\cdot}_F$. We denote $j$-th largest eigenvalue of an operator $A$ as $\lambda_j(A)$ and  $\normOne{A} \coloneqq \sum_{j=1}^{\infty} \abs{\lambda_j(A)}$.  
$I$ stands for an identity operator. 
We also use $\c$ and $\C$ as generalized positive constants, whose values may differ from line to line and depend only on $\kdd$. We use $\asymp$ to denote equality up to a multiplicative constant -- namely, $a_i \asymp b_i$ implies $\c b_i \le a_i \le \C b_i$ for all $i$. We will also write $\Hs$ to denote a RKHS induced by a kernel exhibiting polynomial eigendecay of form $\mu_j \asymp j^{-2s}$ (\ref{polyeigen}).
 
 \section{The algorithm}
 
 GPR attains bias-variance trade-off via imposing a Gaussian Process (GP) prior over the function in question. 
 A GP prior is driven by its mean (typically, assumed to be constant and zero) and covariance function
 \begin{equation}
   f \sim \mathcal{GP}(0, \sigma^2(n\rho)^{-1}k(\cdot,\cdot)),
 \end{equation} 
 where $\rho>0$ is a regularization parameter. In the current study we focus on Matérn kernels, yet the results are also applicable to any covariance function demonstrating polynomial eigendecay and boundness of its eigenfunctions. Posterior distribution over $f$ is also a GP with mean
 \begin{equation}
   \fhat(x) =  k^*(x)\brac{K + n\rho I}^{-1} \boldsymbol{y},
 \end{equation}
 where $\boldsymbol{y}\coloneqq [y_i]_{i=1..n}$, $k^*(x) = [k(x, x_i)]_{i=1..n}$ and $K = [k(x_i,x_j)]_{i,j=1..n}$.

 Alternatively, one can arrive to the same point estimate via Kernel Ridge Regression (KRR) \citep{RW}
 \begin{equation}\label{krr}
   \fhat \coloneqq \arg\max_{f} \cbrac{-\frac{1}{2n}\sum_{i=1}^n \brac{y_i - f(X_i)}^2 - \frac{\rho}{2} \Hknorm{f}^2},
 \end{equation}
 where $\Hknorm{\cdot}$ refers to the RKHS norm, induced by the kernel $k(\cdot,\cdot)$ (see \eqref{rkhsnormdef} for the definition).

 The problem \eqref{krr} is notorious for its high computational complexity being $O(n^3)$, rendering it impossible to scale. As \citep{Zhang2015} suggests, split the set of indices $\{1,2,..,n\}$ into $P$ disjoint sets $\{S_p\}_{p=1}^P$ of size $S\coloneqq\abs{S_p} = n/P$ (we presume $n/P$ is natural for simplicity). Now define $P$ local estimators
 \begin{equation}\label{krrm}
   \fhatm \coloneqq \arg\max_{f} \cbrac{-\frac{1}{2S}\sum_{i\in S_p} \brac{y_i - f(X_i)}^2 - \frac{\rho}{2} \Hknorm{f}^2}
 \end{equation}
 and the averaged one
 \begin{equation}
   \fbar \coloneqq \frac{1}{P}\sum_{p=1}^P \fhatm.
 \end{equation}
 Of course, $P$ cannot grow linearly with $n$, yet for highly smooth classes of functions it can grow as a power of $n$ close to $1$, making the overall complexity $O(n^3/P^2)$ nearly linear (see \eqref{pbound} for details).

 Distribution of $\fbar$ has a complicated nature, while its limiting distribution involves the spectral decomposition of $\kdd$, which is time-consuming to obtain. This significantly complicates the problem of constructing confidence bands for a confidence level $\beta$ of sort
 \begin{equation}\label{band}
   \Prob{\norm{\fbar - \fstar}_2 \le r_{\beta}} =  \beta.
 \end{equation}
  To that end we suggest a non-trivial bootstrap procedure. Classic bootstrap schemes \citep{efron1979} suggest to re-sample the input data. 
  In our case it means solving the problems \eqref{krrm} from scratch for each bootstrap iteration, which is time-consuming. In order to avoid that we suggest to re-sample $\fhatm$ directly, achieving $O(P)$ time complexity.
  Formally, we draw $\fhatmb$ independently and uniformly from $\cbrac{\fhatm}_{p=1}^P$ for $p=1..P$ and define a bootstrap counterpart of the averaged estimator $\fbar$
 \begin{equation}
   \fbarb \coloneqq \frac{1}{P} \sum_{p=1}^P \fhatmb.
 \end{equation}
 Denoting the bootstrap measure as $\mathbb{P}^\flat$ we can now obtain $r_{\beta}^\flat$ as  
 \begin{equation}
   \Probb{\norm{\fbarb - \fbar}_2 \le r_{\beta}^\flat} =  \beta.
 \end{equation}
 We establish closeness of $\mathbb{P}$ and $\mathbb{P}^\flat$ in some sense (see \ref{bootth}), justifying the use of the bootstrap quantile $r_{\beta}^\flat$ instead of the real-world $r_\beta$ in \eqref{band}.
 \begin{remark}
   From a distributed implementation standpoint it may be more convenient and efficient to employ multipliers instead of sampling with return. Namely, the bootstrap counterpart of $\fbar$ can be constructed as $\fbar^w \coloneqq \frac{1}{P} \sum_{p=1}^P u_p \fhatm$ for i.i.d. weights $u_p$ with unit expectation and variance, e. g. $u_p \sim \mathcal{N}(1,1)$. All the results demonstrated for $\fbarb$ are also valid for $\fbar^w$, as we only rely on the first two moments of the bootstrap estimate.
 \end{remark}

\section{Theoretical analysis}
\subsection{Assumptions}
First of all, we impose a polynomial rate of decay on the eigenvalues of $\kdd$.
\begin{assumption}[Polynomial eigendecay]\label{polyeigen}
  Let there exist a constant $s>1/2$ s.t. for  the $j$-th largest eigenvalue $\mu_j$ of $\kdd$
  \begin{equation}
    \mu_j \asymp j^{-2s}.
  \end{equation}
\end{assumption}
As demonstrated in \citep{Yang2017},  \ref{polyeigen} holds for \matern~kernel with smoothness $\alpha$ in a $d$-dimensional space with $s=(2\alpha+d)/2$. Another popular example is a Squared Exponential kernel, which is known to exhibit an exponential rate of eigendecay. With some abuse of formality, our results can be applied in this case with $s=\infty$. Alternatively, the argument can be carefully repeated with minimal augmentation in this case as well.

We also assume the eigenfunctions of the kernel to be bounded. 
The analysis in \citep{Yang2017a,Bhattacharya2017} proves the assumption holds for \matern~kernel under uniform and normal distributions of covariates on a compact.
\begin{assumption}[Boundness of eigenfunctions]\label{eigenbound}
  Denote a normalized eigenfunction corresponding to the $j$-th largest eigenvalue as $\phi_j(\cdot)$.
  Let there exist a positive constant $\phiinf$ s.t. \linebreak $\sup_j\max_{X\in\X} \abs{\phi_j(X)} \le \phiinf $.
\end{assumption}
In conclusion we impose sub-Gaussianity assumption over noise, that being a common relaxation of Gaussianity.
\begin{assumption}[Sub-Gaussianity]\label{subgaussass}
  Let there exist a constant $\g^2$ s.t. for all $a\in \R$
  \begin{equation}
    \E{\exp(a\varepsilon_1)} \le \exp\brac{\frac{\g^2a^2}{2}}.
  \end{equation}
\end{assumption}
\subsection{Theoretical results}
We open the section with a consistency result for $\fhat$. A similar bound is obtained in \citep{Caponnetto2007}. We extend it, explicitly covering undersmoothed priors and providing a slightly tighter bound for that case. Note, it gives a bound in terms of $L_2(\X, \pi)$-norm, which is natural, as an increase of density of $X_i$ in some subset of $\X$ leads to better predictions on the subset. 
\begin{theorem}\label{simplecons}
  Impose \ref{polyeigen}, \ref{eigenbound}, \ref{subgaussass} and let $\fstar \in \Hsnull$ for $\snull \ge s$. Choose $$\rho = n^{-\frac{2s}{2s+1}}.$$ 
  Then for any $\x>1$ and any $\t > 2.6$ on a set of probability at least $1-e^{-\x}-e^{-\t/2}$
  \begin{equation}
    \norm{\fhat-\fstar}_2 \le \C \sqrt{\t\x}\g n^{-\frac{s}{2s+1}} + \C \Hsnullnorm{\fstar} n^{-\frac{\min \{\snull, 2s\} }{2s+1}}
  \end{equation}
  for some $\C>0$ depending only on $s$.
\end{theorem}
The proof (deferred to Appendix \ref{detcons}) relies on the bound, established on a set of high probability. Namely, we define a class of designs we are satisfied with (\ref{designass}), next we demonstrate the measure of the class is high (\ref{lemmadesign}) and establish a consistency result under \ref{designass} (\ref{maintheorem}). 
Here we choose the regularization parameter $\rho$ in a classical manner, acquiring balance between bias and variance in case $\snull = s$.

Under mild assumptions the same minimax-optimal bound may be established for $\fbar$.
\begin{theorem}\label{fbarth}
  Impose \ref{polyeigen}, \ref{eigenbound}, \ref{subgaussass} and let $\fstar \in \Hsnull$, $\snull \ge s$ and 
  \begin{equation}\label{fbarprob}
    P \le \c {\frac{n^{\frac{2s-1}{2s+1}}}{\log n}}.
  \end{equation}
  Choose
  \begin{equation}
    \rho = n^{-\frac{2s}{2s+1}}.
  \end{equation} 
  Then for all $\x>1$ and $\t>2.6$ with probability at least $1-e^{-\x}-e^{-\t/2}$
  \begin{equation}\label{fbarbound}
    \norm{\fbar - \fstar}_2 \le \C \sqrt{\t\x}\g n^{-\frac{s}{2s+1}} + \C \Hsnullnorm{\fstar} n^{-\frac{\min \{\snull, 2s\} }{2s+1}}.
  \end{equation}
\end{theorem}
This theorem is a direct corollary of \ref{fbarlemma}.
The strategy of the proof is to consider Fisher expansion (see \ref{fisher} proven by \cite{Spokoiny2019}) for each $\fhatm$, expressing the discrepancy between the sample-level parameter and its penalized population-level counterpart in terms of Hessian and gradient of the likelihood. 
Next, we bound the Hessian by \ref{lemmadesign}, sum up the expansions and employ additivity of the gradient. Finally, we obtain the concentration via Hanson-Wright inequality.

The expression \eqref{fbarprob} dictates the maximum number of partitions $P$ allowed for the minimax-optimal bound \eqref{fbarbound} to hold. It does indeed match the condition obtained in \citep{Zhang2015}. 

Having obtained the high-probability bound with exponential tail, we can apply integrated tail probability expectation formula to produce the following corollary, repeating the result by \cite{Mucke2018}.
\begin{corollary}
  Impose assumptions of \ref{fbarth}. Then for any positive $\eta$
  \begin{equation}
    \E{\norm{\fbar-\fstar}^\eta_2} = O \brac{n^{-\frac{s\eta}{2s+1}}}.
  \end{equation}
\end{corollary}

Finally, we turn to analysis of the suggested bootstrap scheme. The idea is usual for bootstrap validity results \citep{Chernozhukov2017,chen2018}. First, we establish Gaussian Approximation for the estimator $\fbar$. In order to do so we first notice that by CLT 
\begin{equation}
  \sup_{r>0}\abs{  \Prob{\normSq{\fbar-\ftruerho}_2 < r} -  \Prob{\normSq{\ggamma} < r}} \rightarrow 0
\end{equation}
for $n, P\rightarrow +\infty$, where $ \ftruerho = \E\fbar$ and $\ggamma$ is a centered Gaussian element of a Hilbert space with covariance operator $\Var{\fbar}$. As we are interested in a concentration around $\fstar$ and not $\ftruerho$, we also have to account for the mis-tie between the two, making use of Gaussian Comparison, arriving to
  \begin{equation}\label{boot3}
    \sup_{r>0}\abs{\Prob{\normSq{\fbar-\fstar}_2 < r} -  \Prob{\normSq{\ggamma} < r}} \rightarrow 0.
  \end{equation}
Here we will have to impose undersmoothness of the prior ($\snull > s$) in order to make the remainder term negligible. 

Turning to the bootstrap estimator $\fbarb$, we will use CLT again, which yields
\begin{equation}\label{boot2}
  \sup_{r>0}\abs{\Prob{\normSq{\fbarb-\fbar}_2 < r} -  \Prob{\normSq{\hat\ggamma} < r}} \rightarrow 0
\end{equation}
for a centered Gaussian element of a Hilbert space $\hat\ggamma$ with covariance operator $\Var{\fbarb}$. The final step is to establish closeness of covariance operators of $\ggamma$ and $\hat\ggamma$ and apply Gaussian Comparison obtaining
\begin{equation}\label{boot1}
  \sup_{r>0}\abs{\Prob{\normSq{\ggamma} < r} - \Prob{\normSq{\hat\ggamma} < r}} \rightarrow 0.
\end{equation}
Combining \eqref{boot3}, \eqref{boot2} and \eqref{boot1} will constitute the following claim. 
\begin{theorem}\label{bootth}
  Impose \ref{polyeigen},  \ref{eigenbound}, \ref{subgaussass} and let $\fstar\in\Hsnull$ for $\snull>s$. Choose $$\rho=n^{-\frac{2s}{2s+1}}.$$ Then 
  \begin{equation}
  \begin{split}
    \Rb &\coloneqq \sup_{r>0}\abs{\Prob{\normSq{\fbar - \fstar}_2 < r} - \Probb{\normSq{\fbarb-\fbar}_2 < r}} \\ &\le 
    \C\brac{\frac{\sigma^2n^{\frac{2}{2s+1}}\log^2 n}{{P}}}^{\frac{4s-1}{8s}} + \C n^{-\frac{2\min\cbrac{s,\snull-s}}{2s+1}} \Hsnullnorm{\fstar}^2.
  \end{split}
  \end{equation}
\end{theorem}
The sketched proof is implemented in Appendix \ref{bootproofapp}. 

Naturally, the remainder gets smaller for larger $P$, as it implies a richer set to sample from. On the other hand, \ref{fbarth} imposes an upper bound on $P$. 
Up to logarithmic terms, the choice  of $P=P(n)$ implying both $\Rb = o(1)$ and \eqref{fbarprob} must satisfy 
\begin{equation}\label{pbound}
 n^{\frac{2}{2\snull+1}} \ll P(n) \ll n^{\frac{2\snull-1}{2\snull+1}}
\end{equation}
in order for us to have both high-probability bound and credible bands.
Clearly, it is possible only for $\snull>3/2$. In case of \matern~kernels this translates to $\alpha+d/2 > 3/2$, prohibiting only the case $d=1$ and $\alpha \in (1/2, 1]$.

As discussed in Section \ref{secintro}, choice of an undersmoothed prior is a common way to trade optimality of an estimator for a possibility to construct a confidence interval. But how much do we have to pay? Consider the two summands in the claim of \ref{bootth}. The choice $s = \frac{2}{3} \snull$ implies the former term dominates the latter, hence this choice is the largest reasonable sacrifice. The concentration rate for $\bar{f}$ then would be $n^{-\frac{\snull}{2\snull + 3/2}}$ (instead of the minimax $n^{-\frac{\snull}{2\snull + 1}}$) which is only marginally suboptimal.
\section{Simulation study}
\begin{figure}\centering
  \includegraphics[width=0.6\linewidth]{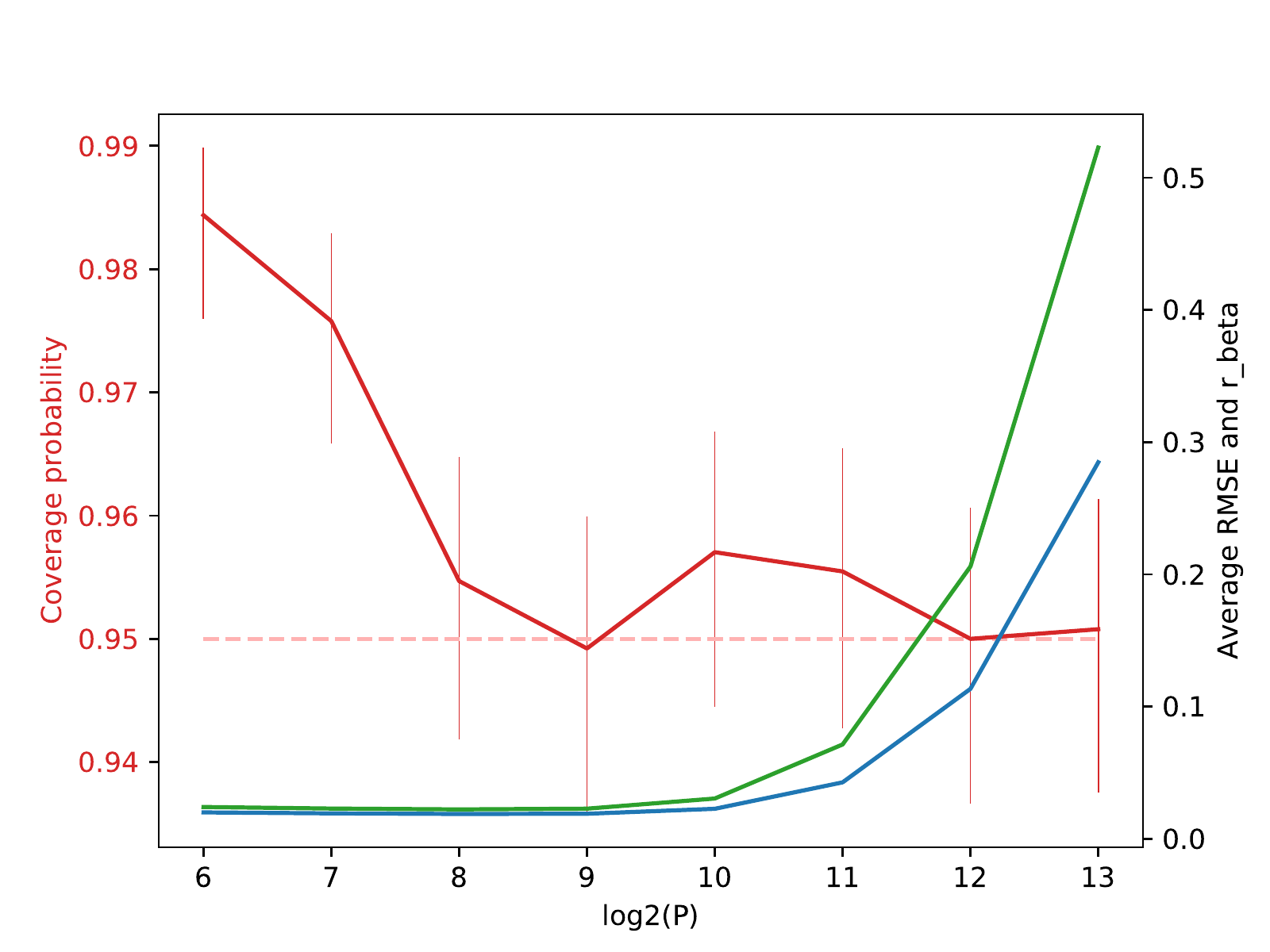}
  \caption{The horizontal axis uses log-2 scale. The nominal significance level is $\beta=0.95$, shown with a dashed line. The estimated coverage probability is shown with a red line, the error bars correspond to  Wilson 95\% point-wise confidence intervals. The green and the blue lines depict the quantile $r_\beta$ (see \eqref{band}) and the average root-mean-square error respectively. Each point of the plot is averaged over $1280$ trials.}
  \label{pic2}
\end{figure}

In this section we study the suggested algorithm experimentally. We choose $\X = [0,1]$ and $\fstar(x) = \sin(\tau x)$, where $\tau \approx 6.28\ldots$ denotes the number of radians in a turn. The design is uniformly and identically sampled from $\X$. The noise $\varepsilon_i$ is Gaussian with variance $\sigma^2=1$, meaning the signal-to-noise ratio is $1/\sqrt{2}$. Nominal confidence level is set to $\beta=0.95$. The number of bootstrap iterations is chosen as $1000$. 
$\kdd$ is chosen to be \matern~kernel with a smoothness index $\alpha=5/2$. 
We choose sample size $n=2^{17}$ and let $P$ vary from $2^6$ to $2^{13}$. The results are shown in Figure~\ref{pic2}. As \ref{bootth} suggests, the number of partitions needs to be large enough and we observe, the method matches the nominal confidence level for $P \ge 2^8$ and does not diverge from it even when excessively large $P$ (e.g. above $2^{10}$) renders the averaged estimator $\fbar$ sub-optimal. The latter effect is described by \ref{fbarth}. Thus, there is a wide range to choose $P$ from, enjoying both minimax optimal estimator $\fbar$ and valid confidence bands.

\section{Conclusion and future work}
The problem of distributed nonparametric regression being of great importance in the light of the ever-growing datasets has earlier received a consistent estimator. 
Namely, the Fast-KRR approach, being an application of divide and conquer paradigm to KRR. 
Its consistency has been demonstrated in terms of risk. 
In this paper we complement these results with a high-probability bound.
Our main contribution is a novel enhancement of the method, providing a confidence band in addition to the point estimate. 
The time complexity of the procedure, being sub-linear in sample size, is dwarfed by the complexity of the Fast-KRR itself, so calculation of the confidence bands is virtually free of charge. 
The theoretical analysis of the procedure is powered by the recent results on Gaussian Comparison \citep{Gotze2019} and a familiar Central Limit Theorem in a Hilbert space \citep{Zalesskii1991}.

Future research should also explore the posteriors of KRRs
\begin{equation}
  f \vert \cbrac{(X_i, y_i)}_{i \in S_p} \propto \exp\cbrac{-\frac{1}{2\sigma^2}\sum_{i=1}^n \brac{y_i - f(X_i)}^2} \exp \cbrac{- \frac{\rho}{2} \Hknorm{f}^2},
\end{equation}
justifying aggregation of both posterior mean and covariance. 
A major obstacle here would be the need to obtain a spectral decomposition of the kernel (as it is involved in the posterior covariance), which is computationally difficult.

There is also a promising alternative. So far all the consistency results for Fast-KRR deal with $L_2$-measure. Obtaining a concentration with respect to a stronger $L_{\infty}$-norm may turn out highly beneficial in the light of the recent research \citep{Yang2017}. There the authors have recognized the posterior covariance of Gaussian Process Regression as an M-estimator and employed the observation to justify its use in construction of sup-norm frequentist confidence sets.

\acks{The research of ``Project Approximative Bayesian inference and model selection for stochastic differential equations (SDEs)'' has been partially funded by Deutsche Forschungsgemeinschaft (DFG) through grant CRC 1294 ``Data Assimilation'', ``Project Approximative Bayesian inference and model selection for stochastic differential equations (SDEs)''.

Further, we would like to thank Vladimir Spokoiny, Evgeniya Sokolova, the three anonymous reviewers and the area chair for the discussions, suggestions, criticism and/or proofreading which have greatly improved the manuscript.
}

\bibliography{main}

\appendix

\section{Proof of $\fhat$ consistency}\label{detcons}
All the claims presented in Appendices \ref{detcons} -- \ref{techapp} implicitly impose \ref{polyeigen}, \ref{eigenbound} and \ref{subgaussass}. We also let $\rho=\rho(n)=o(1)$ and $\fstar \in \Hsnull$. In this section we let $\snull \ge s$.

We open the section with some notation.
Consider the eigenvalues $\cbrac{\mu_j}_{j=1}^\infty$ and normalized eigenfunctions $\cbrac{\phi_j(\cdot)}_{j=1}^\infty$ of $k(\cdot, \cdot)$ w.r.t. continuous measure $\pi$.
Now for $f\in \H$ we have the vector $\f$ of expansion coefficients $\f_j \coloneqq \dotprod{f}{\phi_j}$. Further, we introduce the design matrix $\Phi \in \R^{n\times \infty}$ s.t. $\Phi_{ij} = \phi_j(X_i)$. At this point the estimator \eqref{krr} can be rewritten as 
\begin{equation}
  \fhatbold \coloneqq \arg\max_{\f}\brac{ -\frac{1}{2n} \normSq{\Phi \f - y} - \frac{\rho}{2}\sum_j  \frac{\f_j^2}{\mu_j}}.
\end{equation}
Next, consider a diagonal matrix $M\in \R^{\infty\times\infty}$ s. t. $M_{jj}=\sqrt{\mu_j}$ and define $\param \coloneqq \inv{M}\f$, $\Psi\coloneqq \Phi M$, rewriting \eqref{krr} again
\begin{equation}
  \hatparam \coloneqq \arg\max_{\param}\brac{\underbrace{-\frac{1}{2n}\normSq{\Psi \param - y}}_{L(\param)}-\frac{\rho}{2}\normSq{\param}}.
\end{equation}
We also define 
\begin{equation}
  \trueparam \coloneqq \arg\max_{\param} \E{L(\param)}
\end{equation}
and its penalized counterpart
\begin{equation}\label{thetastarrhodef}
  \trueparamrho \coloneqq \arg\max_{\param} \E{L(\param)} - \frac{\rho}{2}\normSq{\param}.
\end{equation}
Similarly we define $\ftrueboldrho$ and $\fstar_\rho$. 
We also introduce a vector $\eps \in \R^n$ s.t. $\eps_i = \varepsilon_i$.
By the means of trivial calculus we have 
\begin{equation}
  \score \coloneqq \nabla\brac{L(\param) - \mathbb{E}_{\varepsilon}[L(\param)]} = \frac{1}{n} \Psit\eps
\end{equation}
and 
\begin{equation}
  \DrhoSq \coloneqq -\nabla^2\brac{L(\param)-  2\rho\norm{\param}} = \frac{1}{n} \Psit \Psi + \rho I.
\end{equation}

Now we are ready to formulate an assumption we impose on the design $\cbrac{X_i}_{i=1}^n$.

\begin{assumption}[Design regularity]\label{designass}
  Let there exist some positive $\delta$ s. t.
  \begin{equation}
  \norm{\brac{M^2 + \rho I}\minusroot \DrhoSq \brac{M^2 + \rho I}\minusroot - I} \le \delta < 1.
  \end{equation}
\end{assumption}
\ref{designass} can seem to be obscure, but \ref{lemmadesign} guarantees that it holds for a random design with $X_i$ being i.i.d. and distributed w.r.t. a continuous measure.

First, we bound the bias term.
\begin{lemma}\label{biaslemma}
  Let $\fstar \in \Hsnull$ and $\snull \ge s$. Then
\begin{equation}
  \norm{\ftruebold - \ftrueboldrho}^2 \le \rho^{\min\cbrac{\frac{s_\circ}{s},2}}\Hsnullnorm{\fstar}^2.
\end{equation}
\end{lemma}
\begin{proof}
  From the stationarity condition for \eqref{thetastarrhodef} one gets
  \begin{equation}
    \brac{\trueparamrho - \trueparam } = \rho \brac{M^2+\rho I}^{-1}\trueparam
  \end{equation}
  and hence
  \begin{equation}\label{bias2}
    \brac{\ftrueboldrho - \ftruebold } = \rho \brac{M^2+\rho I}^{-1}\ftruebold.
  \end{equation}
  Now using the fact that $\fstar \in \Hsnull$
  \begin{equation}
    \normSq{\rho\brac{M^2+\rho I}^{-1}\ftruebold} \le \C\rho^2 \sum_j \frac{j^{-2\snull}}{\brac{j^{-2s}+\rho}^2}\frac{(\ftruebold_j)^2}{j^{-2\snull}}.
  \end{equation}
  Maximization of $\frac{j^{-2\snull}}{\brac{j^{-2s}+\rho}^2}$ over $j>0$ yields
  \begin{equation}\label{bias1}
      \normSq{\rho\brac{M^2+\rho I}^{-1}\ftruebold} \le \rho^{\min\cbrac{\frac{s_\circ}{s},2}}\Hsnullnorm{\fstar}^2.
  \end{equation}
  Combining \eqref{bias2} and \eqref{bias1} finalizes the proof.
\end{proof}

The next lemma bounds the right-hand side of Fisher expansion (\ref{fisher}). 

\begin{lemma}\label{scoreconcentration}
  For all $\x>0$
  \begin{equation}
    \Prob{\norm{\DrhoInv\score} \le \C\g\sqrt{\frac{{1+\sqrt{\x}+2\x}}{\rho^{1/(2s)}n(1-\delta)} }} \ge 1-e^{-\x}.
   \end{equation}
\end{lemma}
\begin{proof}
  By the definition of $\score$ 
  \begin{equation}
    \normSq{\DrhoInv\score} = \frac{1}{n^2} \epst\Psit \brac{\frac{1}{n}\Psit\Psi + \rho I}^{-1} \Psi \eps.
  \end{equation}
  Clearly, 
  \begin{equation}
    \brac{\frac{1}{n}\Psit\Psi + \rho I} \ge (1-\delta) \brac{M^2 + \rho I} 
  \end{equation}
  and hence 
  \begin{equation}
    \normSq{\DrhoInv\score} \le \epst {\frac{1}{n^2}\Psi^T \inv{\brac{(1-\delta) \brac{M^2 + \rho I} }} \Psi} \eps.
  \end{equation}
  Applying \ref{scoreconcentrationGeneral} completes the argument.
\end{proof}

Now we are ready to establish a consistency result. 

\begin{lemma}\label{maintheorem}
  Impose \ref{designass}. Then on a set of probability at least $1-e^{-\x}$ for all $\x>1$
  \begin{equation}
    \norm{\fhat-\fstar}_2 \le  \C  \sqrt{\underbrace{\frac{\g^2{\x}}{(1-\delta)\rho^{1/(2s)}n}}_{Variance}+\underbrace{\rho^{\min\cbrac{\frac{s_\circ}{s},2}}\Hsnullnorm{\fstar}^2}_{Bias}}.
  \end{equation}
\end{lemma}
\begin{proof}
  We apply \ref{scoreconcentration} along with \ref{fisher}, which yield for some positive $\C$ on a set of probability at least $1-e^{-\x}$ for any positive $\x$
  \begin{equation}
    \norm{\Drho\brac{\hatparam - \trueparamrho}} \le \C\g\sqrt{\frac{{1+\sqrt{\x}+2\x}}{\rho^{1/(2s)}n(1-\delta)} }.
  \end{equation}
  But clearly, 
  \begin{equation}
    (1-\delta)\normSq{\fhatbold-\ftrueboldrho}_2 = \normSq{\sqrt{1-\delta}M(\hatparam-\trueparam)} \le (1-\delta) \norm{\Drho\brac{\hatparam - \trueparam}}^2.
  \end{equation}
  In order to bound the bias term we apply \ref{biaslemma}, constituting the claim.
\end{proof}
The proof of \ref{simplecons} is almost trivial now.
\begin{proof}[Proof of \ref{simplecons}]
The proof consists in applying \ref{lemmadesign} followed by applying \ref{maintheorem}.
\end{proof}

\section{Proof of $\bar f$ consistency}\label{fbarsec}
Denote a map from $\X$ to $\R^{\infty}$
\begin{equation}
  \psi(X) \coloneqq \brac{\sqrt{\mu_1}\phi_1(X)~~ \sqrt{\mu_2}\phi_2(X)~~ \sqrt{\mu_3}\phi_3(X)~~ ... }^T \in \R^{\infty}
\end{equation}
and  $\psi_i \coloneqq  \psi(X_i)$.

We have earlier introduced the objects related to the global estimator $\fhat$, such as $\Psi$, $\DrhoSq$, $\hatparam$, $\fhatbold$, $\eps$. In this section we make use of their local counterparts related to $\fhatm$ such as $\Psim$, $\DrhomSq$, $\hatparamm$, $\fhatboldm$, $\epsm$. 
Also define $\barparam \coloneqq \frac{1}{P} \sum_p \hatparamm$ and $\fbarbold \coloneqq \frac{1}{P}\sum_p \fhatboldm$. Throughout the section we let $\snull \ge s$. 
\begin{lemma}\label{fbarlemma}
  Let
  \begin{equation}\label{fbarprobLemma}
    P \le \c {\frac{n^{\frac{2s-1}{2s+1}}}{\log n}}.
  \end{equation}
  Choose
  \begin{equation}
    \rho = n^{-\frac{2s}{2s+1}}.
  \end{equation} 
  Then for all $\x>0$ and $\t>0$ with probability at least $1-e^{-\x}-\brac{e^{\t}-\t-1}^{-1}$
  \begin{equation}\label{fbarbound}
    \norm{\fbar - \fstar}_2 \le \C \sqrt{\t\brac{1+2\x+\sqrt{\x}}}\g n^{-\frac{s}{2s+1}} + \C \Hsnullnorm{\fstar} n^{-\frac{\min \{\snull, 2s\} }{2s+1}}.
  \end{equation}
\end{lemma}
\begin{proof}
  Using Fisher expansion (\ref{fisher})
  \begin{equation}
  \begin{split}
    \Drhom\brac{\hatparamm - \trueparamrho} &= \frac{1}{S} \DrhomInv \Psim \epsm \\
    &= \frac{1}{S} \DrhomInv \brac{\sum_{i \in S_p} \psi_i \varepsilon_i}.
  \end{split}
  \end{equation}
  Now by \ref{lemmadesign} with probability at least $1-\brac{e^{\t}-\t-1}^{-1}$ we have
  \begin{equation}
  \begin{split}
    \norm{\brac{1-\delta(S, \t)}\proot\brac{M^2+\rho I}\proot \brac{\barparam - \trueparamrho}}  &\le \norm{\frac{1}{P} \sum_{p=1}^P\Drhom\brac{\hatparamm - \trueparamrho}} \\ &\le \norm{\frac{1}{n} \brac{1-\delta(S, \t)}\minusroot \brac{M^2+\rho I}\minusroot \Psi \eps}
  \end{split}
  \end{equation}
  and hence
  \begin{equation}
    \norm{\fbarbold - \ftrueboldrho} \le \frac{1}{1-\delta(S,\t)} \norm{\frac{1}{n} \brac{M^2+\rho I}\minusroot \Psi \eps}.
  \end{equation}
  Next apply \ref{scoreconcentrationGeneral}. On a set of probability at least $1-e^{-\x}$ for all $\x>0$ we have
  \begin{equation}
    \norm{\fbarbold - \ftrueboldrho} \le \frac{ \C\g}{1-\delta(S,\t)}\sqrt{\frac{{1+\sqrt{\x}+2\x}}{\rho^{1/(2s)}n}}.
  \end{equation}
  Assumption \eqref{fbarprobLemma} implies $1-\delta(S, \t) > 1/2$.
  The bias term is controlled by \ref{biaslemma}.
\end{proof}

\section{Bootstrap validity proof}\label{bootproofapp}
Define $P$ i.i.d. vectors $\gboldm \coloneqq \fhatboldm - \ftrueboldrho$, denote $\Sigma = \E{\gboldm\gboldm^T}$ and 
 $\hat \Sigma\coloneqq \frac{1}{P}\sum_p{\gboldm\gboldm^T}$.
Throughout the section we let $\snull > s$. 

 \begin{lemma}\label{fbargaragain}
   Consider a centered Gaussian element of a Hilbert space $\ggamma$ with a covariance operator $\Sigma$.
  Then for all positive $\t > 2.6$ and $\delta=\delta(S, \t)$ coming from \ref{lemmadesign}
  \begin{equation}
    (1+\delta)^{-1} \Sigmastar \le n\Var{\fbarbold - \ftrueboldrho} \le (1-\delta)^{-1} \Sigmastar,
  \end{equation}
  \begin{equation}
    (1+\delta)^{-1} \Sigmastar\le  {S}\Var{\fhatboldm - \ftrueboldrho} \le (1-\delta)^{-1}\Sigmastar
  \end{equation}
  and
  \begin{equation}
    \tr{\Sigma} \le\C \sigma^2 S^{-1} \rho^{-1/(2s)},
  \end{equation}
  where $\Sigmastar \coloneqq \sigma^2\brac{M^2+\rho I}^{-2} M^4$.
  Morover, uniformly for positive $r$
   \begin{equation}
   \begin{split}
     &\abs{\Prob{ \norm{\fbar - \fstar}_2\le r} - \Prob{P^{-1/2}\norm{\ggamma}\le r}} \\&\le \C P^{-1/2} + \C  n^{-\frac{2\min\cbrac{s,\snull-s}}{2s+1}} \Hsnullnorm{\fstar}^2.
   \end{split}
   \end{equation}
 \end{lemma}
 \begin{proof}
   Using Fisher expansion (\ref{fisher}) we have
   \begin{equation}
     \brac{\hatparamm - \trueparamrho} = \frac{1}{S} \DrhomSqInv \brac{\sum_{i\in S_p} \psi_i \varepsilon_i}
   \end{equation}
   and hence by \ref{lemmadesign} we have
   \begin{equation}
     M\brac{\barparam - \trueparamrho} \ge \frac{1}{n(1+\delta)} M\inv{\brac{M^2+\rho I}}\brac{\sum_{i} \psi_i \varepsilon_i}.
   \end{equation}
   In the same way we obtain the less-or-equal inequality constituting the first part of the claim.  The bound for $\tr \Sigma$ is obtained by \ref{effdimlemma}
   \begin{equation}
     \tr{S\Sigma} \le \C\sigma^2(1-\delta)^{-1} \sum_j \frac{\mu^2_j}{\brac{\mu_j+\rho}^2} \le \C\sigma^2 \rho^{-1/(2s)}. 
   \end{equation}
   This also bounds the six largest eigenvalues of $S\Sigma$ away from zero. 
   
   Next we use \ref{scoreconcentration}, \ref{lemmadesign} and \ref{fisher} in the same way we did in the proof of \ref{maintheorem} and have for an arbitrary $p$ and all positive $\x$
   \begin{equation}
     \Prob{\norm{\gboldm} \ge \C \g \sqrt{\frac{\x}{\rho^{1/(2s)}S}}} \le e^{-\x}.
   \end{equation}
   Hence, using integrated tail probability expectation formula we have
   \begin{equation}
     \E{\norm{\gboldm}^3} \le \brac{\frac{\C^2 \g^2}{\rho^{1/(2s)}S}}^{3/2}.
   \end{equation} 
   By \ref{effdimlemma} 
   \begin{equation}
     \E{\norm{\gboldm}^2} \asymp \frac{\C^2 \g^2}{\rho^{1/(2s)}S}.
   \end{equation} 
  Now we are ready to apply \ref{cltlemma} which yields
   \begin{equation}
   \begin{split}
     \abs{\Prob{ \norm{\fbar - \fstar_\rho}_2\le r} - \Prob{P^{-1/2}\norm{\ggamma}\le r}} \le \C  \frac{\g^3}{\sigma^3}P^{-1/2} .
   \end{split}
   \end{equation}
   The last step is to apply \ref{gcomp} accounting for the mis-tie between $\fstar$ and $\fstar_\rho$, which is controlled by \ref{biaslemma}.
 \end{proof}

\begin{lemma}\label{sigmasigmahatlemma}
  For all $\t > 2.6$ on a set of probability at least $1-e^{-\t/2}-P^{-3}$ 
  \begin{equation}
    S\norm{\hat\Sigma-\Sigma} \le \Delta(P, \t) \coloneqq  \C\sigma^2 \sqrt{\t}\rho^{-1/(2s)}P^{-1/2}\log P.
  \end{equation}
\end{lemma}
\begin{proof}
   Consider $P$ i.i.d. matrices
 \begin{equation}
   \Omega_p  = \gboldm\gboldm^T - \Sigma,
 \end{equation}
   \ref{fbargaragain} yields
  \begin{equation}
      (1+\delta)^{-1} M^4 {\brac{M^2+\rho I}}^{-2} \le \Sigma\le (1-\delta)^{-1} M^4{\brac{M^2+\rho I}}^{-2}.
  \end{equation}
  On a set of probability at least $1-P^{-3}$ for all $p$ we have 
  \begin{equation}
    \tr{\gboldm\gboldm^T} \le \C \frac{\g^2 \log P}{\rho^{1/(2s)}S},
  \end{equation}
  at the same time by \ref{fbargaragain}
  \begin{equation}
    \tr{\Sigma}  \le\C \sigma^2\rho^{-1/(2s)}S^{-1}
  \end{equation}
  and hence 
  \begin{equation}
    \tr{\Omega_p} \le \C \sigma^2\rho^{-1/(2s)}S^{-1}\log P.
  \end{equation}
  Clearly, 
  \begin{equation}
    \norm{\Omega_p^2} \le \tr{\Omega_p^2} \le C \sigma^2\rho^{-1/s}S^{-2}\log^2 P.
  \end{equation}
  The rest is due to \ref{bernmat}.
\end{proof}

\begin{lemma}\label{sigmanormonelemma}
  On the same set which the claim of \ref{sigmasigmahatlemma} holds on (of probability at least $1-e^{-\t/2}-P^{-3}$) for an arbitrary $\t>2.6$
  \begin{equation}
    S\normOne{\Sigma-\hat\Sigma} \le \C\frac{\Delta(P,\t)^{1-1/(4s)}}{\rho^{1/(2s)}}.
  \end{equation}
\end{lemma}
\begin{proof}
  Denote $\Delta \coloneqq \Delta(P, \t)$.
  \begin{equation}
  \begin{split}
    S\normOne{{\Sigma - \hat\Sigma}} &= S\brac{\sum_{j^{2s} \le 1/(\rho\sqrt{\Delta})} + \sum_{j^{2s} > 1/(\rho\sqrt{\Delta})}} \abs{\lambda_j(\Sigma - \hat\Sigma)} \\
    &\le \C\frac{\Delta^{1-1/(4s)}}{\rho^{1/(2s)}} + \frac{\C}{\rho^2} \int_{u^{2s} > 1/(\rho\sqrt{\Delta})} \frac{du}{u^4s} \\
    & \le \C\frac{\Delta^{1-1/(4s)}}{\rho^{1/(2s)}} .
  \end{split}
  \end{equation}
\end{proof}

\begin{lemma}\label{fbarboldbgarlemma}
   Consider $\ggammab\sim \mathcal{N}(0,\hat\Sigma)$. Then uniformly for positive $r$
\begin{equation}
  \abs{\Probb{\norm{\fbarboldb - \fbarbold} < r } - \Probb{P^{-1/2} \norm{\ggammab}<r}} \le \C \frac{\g^3}{\sigma^3}P^{-1/2}.
\end{equation}
\end{lemma}
\begin{proof}
The proof consists in applying CLT. In order to estimate the moments involved in its residual term we apply \ref{simplecons} to the local estimates $\fhatm$ and obtain on a set of probability at least $1-Pe^{-\x} - Pe^{-\t/2}$ (choosing $2\x=\t=4\log P$) for $S$ large enough
\begin{equation}
  \norm{\gboldm} \le 1,
\end{equation}
which enables us to apply Hoeffding's inequality, yielding concentration for $\u$ chosen as $\sqrt{\log P}$
\begin{equation}
  \Prob{\abs{\Eb{\normSq{\gboldm}} -  \E{\normSq{\gboldm}}} > \u P^{-1/2} } \le 2e^{-2 \u^2}
\end{equation}
and also
\begin{equation}
  \Prob{\abs{\Eb{\norm{\gboldm}^3} -  \E{\norm{\gboldm}^3}} > \u P^{-1/2} } \le 2e^{-2 \u^2}.
\end{equation}
Now we can bound the moments involved in \ref{cltlemma}
  \begin{equation}
    \Eb{\norm{\fhatboldmb - \fbarbold}^2} \asymp \E{\norm{\fhatboldm - \fstarbold}^2} \asymp \brac{\sigma\rho^{-1/(2s)}}^2,
  \end{equation}
  \begin{equation}
    \Eb{\norm{\fhatboldmb - \fbarbold}^3} \asymp \E{\norm{\fhatboldm - \fstarbold}^3} \asymp \brac{\g\rho^{-1/(2s)}}^3.
  \end{equation}
  \ref{sigmasigmahatlemma} (choose $\t = 2\log P$) demonstrates boundness of the six largest eigenvalues of $\hat\Sigma$ away from zero. Finally, we use \ref{condlemma} to account for the conditioning (probability of the set is at least $1-1/P$).
\end{proof}

\begin{proof}[Proof of \ref{bootth}]
  Finally, we are in position to demonstrate closeness of $\mathbb{P}$ and $\mathbb{P}^\flat$. We apply \ref{fbarboldbgarlemma} and \ref{fbargaragain}, obtaining two Gaussian approximations and compare them by the means of \ref{gcomp}.
  We use \ref{sigmanormonelemma} (choose $\t=2\log P$) to establish closeness of the covariance operators of the limiting distributions. Finally, account for conditioning (\ref{condlemma}). 
\end{proof}
\section{Technical Results}\label{techapp}
The following lemma aids to bound the right-hand side of Fisher expansion (\ref{fisher}). 
\begin{lemma}
 For all positive $\x$
 \begin{equation}\label{scoreconcentrationGeneral}
   \Prob{\norm{\frac{1}{n}\brac{M^2+\rho I}\minusroot \Psi \eps} \le \C\g\sqrt{\frac{{1+\sqrt{\x}+2\x}}{\rho^{1/(2s)}n}}} \ge 1 - e^{-\x}.
 \end{equation}
\end{lemma}
\begin{proof}
  Trivially,
  \begin{equation}
    \norm{\frac{1}{n}\brac{M^2+\rho I}\minusroot \Psi \eps}^2 =  \norm{\eps^T \underbrace{\frac{1}{n^2} \Psi^T \brac{M^2+\rho I}^{-1} \Psi}_A \eps}.
  \end{equation}
  Now we bound every diagonal element of the matrix $A$.
  \begin{equation}
  \begin{split}
    \max_iA_{ii} &\le \frac{\phiinf^2}{n^2}\sum_{j=1}^\infty  \frac{\mu_j}{\mu_j+\rho}.
  \end{split}
  \end{equation}
  Now employ \ref{effdimlemma}, which holds due to \ref{polyeigen}, and obtain
  \begin{equation}
    \max_i A_{ii} \le \C\rho^{-1/(2s)}{n^{-2}}.
  \end{equation}
  Therefore, 
  \begin{equation}
    \tr{A} \le  {\C\rho^{-1/(2s)}}{n^{-1}}
  \end{equation}
  and trivially
  \begin{equation}
    \tr{A^2} \le \brac{{\C\rho^{-1/(2s)}}{n^{-1}}}^2,
  \end{equation}
  \begin{equation}
    \normF{A} = \sqrt{\tr{A^TA}} \le {\C\rho^{-1/(2s)}}{n^{-1}}.
  \end{equation}
  Finally, we are ready to employ Hanson-Wright inequality \citep{rudelson2013}, constituting the claim.
\end{proof}

Below we demonstrate that \ref{designass} holds with high probability under general assumptions.
\begin{lemma}\label{lemmadesign}
  Let $\cbrac{\phi_j}$ and $\cbrac{\mu_j}$ be eigenfunctions and eigenvalues of $k(\cdot,\cdot)$ w.r.t. $\pi$.
  Then \ref{designass} holds for some $\C > 0$ and arbitrary $\t>2.6$ with 
  \begin{equation}
    \delta =  \C \rho^{-1/(2s)}\brac{\sqrt{\frac{\t}{n}} + \frac{\t}{n}}
  \end{equation}
  on a set of probability at least $1-e^{-\t/2}$.
\end{lemma}
\begin{proof}
  Consider matrices $\Psi^i \in \R^{\infty\times\infty}$ s. t. $\Psi^i_{jk} = \sqrt{\mu_j\mu_k}\phi_j(X_i) \phi_k(X_i)$. Denote 
  \begin{equation}
    \Omega_i = \brac{M^2+\rho I}\minusroot\brac{\Psi^i + \rho I}\brac{M^2+\rho I}\minusroot - I.
  \end{equation}
  Observe
  \begin{equation}
    \frac{1}{n}\sum_i^n \Omega_i= \brac{M^2 + \rho I}\minusroot \brac{\frac{1}{n}\Psit \Psi + \rho I} \brac{M^2 + \rho I}\minusroot - I.
  \end{equation}
  Due to the fact that the eigenfunctions are normalized, $\E{\Omega_i} = 0$. 
  
  Below $j$ and $k$, being summation indexes, always run from $1$ to $\infty$ unless specified otherwise. 
  Clearly the maximum eigenvalue of a p.s.d. matrix does not exceed its trace. Hence, using \ref{effdimlemma}
  \begin{equation}
    \begin{split}
    \norm{\brac{M^2+\rho I}\minusroot\Psi^i\brac{M^2+\rho I}\minusroot} &\le \C \sum_j \frac{\mu_j}{\mu_j+\rho}\\
    &\le \C \rho^{-1/(2s)}
    \end{split}
  \end{equation}
  and 
  \begin{equation}
  \begin{split}
    \norm{\brac{M^2+\rho I}\minusroot \rho I  \brac{M^2+\rho I}\minusroot  } 
    &\le \rho \max_j \frac{1}{\mu_j + \rho } \\ 
    &\le 1.
  \end{split}
  \end{equation}
  Further, using the choice of $\rho$ we have
  \begin{equation}
  \begin{split}
    \norm{\Omega_i} &\le \C \rho^{-1/(2s)} + 2\\
    & \le \C \rho^{-1/(2s)}.
  \end{split}
  \end{equation}
  Now the goal is to bound $\tr{\E{\Omega_i^2}}$. First, we observe
  \begin{equation}
    \E{\brac{M^2 + \rho I}\minusroot (\Psi^i+\rho I) \brac{M^2 + \rho I}\minusroot} = I 
  \end{equation}  
  due to the fact that $\E{\Psi^i} = M^2$.
  Therefore, 
  \begin{equation}
    {\E{\Omega_i^2}} = \E{\underbrace{\brac{\Frho\minusroot \brac{\Psi^i + \rho I} \Frho\minusroot}^2}_A} - I.
  \end{equation}
  For its trace, using $\mathbb{I}[\cdot]$ to denote an indicator, we have 
  \begin{equation}
  \begin{split}
     \tr{\E{\Omega_i^2}} &\le \sum_j \brac{\sum_k {\frac{\brac{\phiinf^2\sqrt{\mu_j\mu_k} + \rho\indicator{j=k}}^2}{\brac{\mu_j + \rho}\brac{\mu_k+\rho}}} -1} \\
    &\le \sum_j\sum_k {\frac{\brac{\phiinf^2\sqrt{\mu_j\mu_k}}^2}{\brac{\mu_j + \rho}\brac{\mu_k+\rho}}} + \abs{\sum_j \brac{\brac{\frac{\phiinf^2 \mu_j+\rho}{\mu_j+\rho}}^2-1}}\\
    &\eqqcolon T_1 + T_2.
  \end{split}
  \end{equation}
  Using \ref{effdimlemma} twice, relying on the choice of $\rho$ we have
  \begin{equation}
  \begin{split}
    T_1 &\le \C \sum_j \brac{\frac{\mu_j}{\mu_j+\rho}\sum_k \frac{\mu_k}{\mu_k+\rho}} \\
    & \le \C \sum_j \frac{\mu_j}{\mu_j+\rho} \times \rho^{-1/(2s)}\\
    & \le \C {\rho^{-1/s}}.
  \end{split}
  \end{equation}
  The treatment of the second term uses decay of $\mu_j$
  \begin{equation}
  \begin{split}
    T_2 &\le \C \abs{\sum_j \frac{\mu_j^2+\rho\mu_j}{(\mu_j+\rho \mu_j)^2}} \\
    & \le \C\sum_j \frac{\rho\mu_j}{\mu_j^2+2\mu_j\rho+\rho^2}\\
    & \le \C \sum_j\frac{\mu_j}{\mu_j + \rho}\\
    & \le \C\rho^{-1/(2s)},
  \end{split}
  \end{equation}
  where \ref{effdimlemma} was used again. Therefore, we have 
  \begin{equation}
    \norm{\E{\Omega_i^2}} \le \tr{\E{\Omega_i^2}} \le \C \rho^{-1/s}.
  \end{equation}
  Finally, apply \ref{bernmat}, demonstrating that \ref{designass} holds for 
  \begin{equation}
  \begin{split}
    \delta =  \C \rho^{-1/(2s)}\brac{\sqrt{\frac{\t}{n}} + \frac{\t}{n}}
  \end{split}
  \end{equation}
  with probability at least $1-e^{-\t/2}$.
\end{proof}
The next lemma, being almost folklore (see \cite{Zhang2015,Yang2017,Spokoiny2019}), bounds the effective dimensionality.
\begin{lemma}\label{effdimlemma}
  \begin{equation}
    \sum_{j=1}^\infty \frac{\mu_j}{\mu_j + \rho} \asymp \rho^{-1/(2s)}.
  \end{equation}
\end{lemma}
\begin{proof}
  \begin{equation}
  \begin{split}
    \sum_{j=1}^\infty \frac{\mu_j}{\mu_j + \rho} \asymp \underbrace{\sum_{j^{2s} \le 1/\rho} \frac{1}{1 + j^{2s}\rho}}_{T_1} + \underbrace{\sum_{j^{2s} > 1/\rho} \frac{1}{1 + j^{2s}\rho}}_{T_2}.
  \end{split}
  \end{equation}
  \begin{equation}
  \begin{split}
    T_1 \asymp \rho^{-1/(2s)}.
  \end{split}
  \end{equation}
  \begin{equation}
  \begin{split}
    T_2 &= \frac{1}{\rho}\sum_{j^{2s} > 1/\rho} \frac{1}{1/\rho + j^{2s}} \\
    & \asymp \frac{1}{\rho}\sum_{j^{2s} > 1/\rho} \frac{1}{j^{2s}}\\
    & \asymp \frac{1}{\rho} \int_{u = \rho^{-1/(2s)}}^{+\infty} \frac{du}{u^{2s}}\\
    &\asymp \rho^{-1/2s}.
  \end{split}
  \end{equation}
\end{proof}
The next lemma quantifies how much a measure of a set changes after conditioning. 
\begin{lemma}\label{condlemma}
  Consider a measure $\mathbb{P}$ and two measurable sets $A$ and $B$. Then 
  \begin{equation}
    \abs{\Prob{A} - \Prob{A|B}} \le 2 \Prob{\bar B}.
  \end{equation}
\end{lemma}
\begin{proof}
  \begin{equation}
  \begin{split}
    \abs{\Prob{A} - \Prob{A|B}} &= \abs{\Prob{A|B}\Prob{B} + \Prob{A|\bar B}\Prob{\bar B} - \Prob{A|B} } \\
    &= \abs{\Prob{A|B}(\Prob{B}-1) + \Prob{A|\bar B}\Prob{\bar B} } \\
    &\le 2\Prob{\bar B}.
  \end{split}
  \end{equation}
\end{proof}
\section{Tools}
This section briefly cites the results we relied upon.
\subsection{Consistency of penalized maximum likelihood estimation}
Consider a quadratic concave likelihood $L(\param) = L(\param, Y)$ for $\param$ being an infinite-dimensional parameter and $Y$ denoting the random data. 
The following result quantifies the mis-tie between the penalized  sample-level estimate
\begin{equation}
  \hatparam \coloneqq \arg\max_{\param} \underbrace{L(\param) - \rho \norm{\param}^2}_{\Lrho(\param)},
\end{equation}
penalized population-level estimate 
\begin{equation}
  \trueparamrho \coloneqq \arg\max_{\param} \E{L(\param)} - \rho \norm{\param}^2
\end{equation}
and non-penalized population-level estimate
\begin{equation}
  \trueparam \coloneqq \arg\max_{\param} \E{L(\param)}.
\end{equation}
Also define 
\begin{equation}
  \DrhoSq \coloneqq -\nabla^2\brac{\E{L(\param)} - \rho \normSq{\param}} \text{ and } \nabla \zeta \coloneqq \nabla\zeta(\param) = \nabla \brac{L(\param) - \E{L(\param)}}.
\end{equation}
\begin{lemma}\label{fisher}
    Fisher expansion holds:
  \begin{equation}
    \Drho \brac{\hatparam - \trueparamrho} = \DrhoInv \nabla \zeta.
  \end{equation}
\end{lemma}
\begin{proof}Using Taylor expansion around the stationary point $\hatparam$ we have
  \begin{equation}
    \nabla\Lrho(\param)  =  -\DrhoSq \brac{\param - \hatparam}.
  \end{equation}
  Now we notice $\nabla \E{\Lrho(\trueparamrho)} = 0$ and obtain
  \begin{equation}
    \nabla \Lrho(\trueparamrho) = \nabla \zeta = -\DrhoSq \brac{\trueparamrho - \hatparam}.
  \end{equation}
  Finally, relying on the fact that $\rho>0$, thus $\DrhoSq$ is invertible, we multiply both sides of the last equation by $\DrhoInv$.
\end{proof}
This result has been generalized in \cite{Spokoiny2019} (Theorem 2.2).

\subsection{Gaussian Comparison}
\begin{lemma}[Theorem 2.1 by \cite{Gotze2019}]\label{gcomp}
  Consider two centered Gaussian elements of a Hilbert space $\zeta$ and $\eta$ with covariance operators $\Sigma_1$ and $\Sigma_2$ and a deterministic Hilbert element $a$. Denote
  \begin{equation}
    \kappa(\Sigma) = \brac{\sqrt{\sum_{j>1}\lambda_j^2(\Sigma)} \sqrt{\sum_{j\ge 1} \lambda_j^2(\Sigma)}}^{-1/2}.
  \end{equation} 
  Then
  \begin{equation}
    \sup_{r>0} \abs{\Prob{\norm{\zeta - a} < r} - \Prob{\norm{\eta} < r}} \le \C \brac{\kappa(\Sigma_1) + \kappa(\Sigma_2)}\brac{\normOne{\Sigma_1 - \Sigma_2} + \normSq{a}}.
  \end{equation}
\end{lemma}

\subsection{Central Limit Theorem}
\begin{lemma}[Corollary of the main theorem by \cite{Zalesskii1991}]\label{cltlemma}
  Consider centered $X_1$, $X_2$, .., $X_n$ being i.i.d. elements of Hilbert space with covariance operator $V$ and a centered Gaussian element $Y$ with the same covariance operator.
  Then
  \begin{equation}
    \sup_{r>0} \abs{\Prob{\norm{n^{-1/2}\sum_{i=1}^n X_i}<r} - \Prob{\norm{Y} < r}} \le R,
  \end{equation}
  where 
  \begin{equation}
    R \coloneqq \C \brac{\prod_{i=1}^6 \lambda_i(V)}^{-1} \E{\normSq{X_1}}^{-3/2}\E{\norm{X_1}^3} n^{-1/2}.
  \end{equation}
\end{lemma}

\subsection{Bernstein matrix inequality}
\begin{lemma}[Corollary of Theorem 3.3 \cite{hsu2012}]\label{bernmat}
  Consider centered i.i.d. matrices $X_1$, $X_2$, .., $X_n$.
  Let for some positive $A$ and $B$
  \begin{equation}
    \lambda_1(X_i) \le A,
  \end{equation}
  \begin{equation}
    \lambda_1 \brac{\frac{1}{n} \sum_{i=1}^n \E{X_i^2} } \le B ,
  \end{equation}
  \begin{equation}
    \tr{\frac{1}{n} \sum_{i=1}^n \E{X_i^2} } \le B .
  \end{equation}
  Then for any $\t > 0$
  \begin{equation}
    \Prob{\lambda_1\brac{\frac{1}{n}\sum_{i=1}^n X_i} > \sqrt{\frac{2B\t}{n}} + \frac{At}{3n}} \le \t\brac{e^{\t}-\t-1}^{-1}.
  \end{equation}
\end{lemma}
Note, for positive $\t\ge 2.6$ 
\begin{equation}
  \t\brac{e^{\t}-\t-1}^{-1} \le e^{-\t/2}.
\end{equation}

\end{document}